\documentclass[11pt]{article}

\makeatletter
    \renewcommand\section{\@startsection {section}{1}{\z@}%
                                       {-3.5ex \@plus -1ex \@minus -.2ex}%
                                       {2.3ex \@plus.2ex}%
                                       {\normalfont\fontfamily{phv}\fontsize{16}{19}\bfseries}}
    \renewcommand\subsection{\@startsection{subsection}{2}{\z@}%
                                         {-3.25ex\@plus -1ex \@minus -.2ex}%
                                         {1.5ex \@plus .2ex}%
                                         {\normalfont\fontfamily{phv}\fontsize{14}{17}\bfseries}}
    \renewcommand\subsubsection{\@startsection{subsubsection}{3}{\z@}%
                                        {-3.25ex\@plus -1ex \@minus -.2ex}%
                                         {1.5ex \@plus .2ex}%
                                         {\normalfont\normalsize\fontfamily{phv}\fontsize{14}{17}\selectfont}}
    \makeatother
\usepackage{amsmath}
\usepackage{graphicx}
\usepackage{enumerate}
\usepackage{url}
\usepackage{cite}
\usepackage{textcomp}
\usepackage{xcolor}
\usepackage{float}
\usepackage{wrapfig}
\usepackage{algpseudocode}
\usepackage{multirow}
\usepackage{amsfonts}
\usepackage{mathrsfs}
\usepackage{amssymb,latexsym}
\usepackage{commath}
\usepackage{caption}
\usepackage{subcaption}
\usepackage{booktabs}
\usepackage{mathtools}
\usepackage{todonotes}
\usepackage{soul}
\usepackage[ruled,vlined]{algorithm2e}
\usepackage{multicol}
\usepackage{amsbsy}
\usepackage{comment}
\usepackage[T1]{fontenc}
\usepackage[cmintegrals]{newtxmath}
\usepackage{bm}
\usepackage{array}
\usepackage{longtable}
\usepackage{setspace}
\usepackage{hyperref}
\hypersetup{
	linkcolor=cyan,
	filecolor=magenta,
	urlcolor=blue,
}
\newcommand{\vect}[1]{\boldsymbol{#1}}
\graphicspath{{Figs/}}

\begin{document}

  \def\spacingset#1{\renewcommand{\baselinestretch}{#1}\small\normalsize} \spacingset{1}
          
  \title{Geometry-aware Active Learning of Spatiotemporal Dynamic Systems}

  \author{Xizhuo (Cici) Zhang and Bing Yao\footnote{Corresponding author: Bing Yao, byao3@utk.edu}  \\  Department of Industrial \& Systems Engineering, The University of Tennessee, \\ Knoxville, TN,  USA, 37996 }

  \maketitle

  \begin{abstract}
  Rapid developments in advanced sensing and imaging have significantly enhanced information visibility, opening opportunities for predictive modeling of complex dynamic systems. However, sensing signals acquired from such complex systems are often distributed across 3D geometries and rapidly evolving over time, posing significant challenges in spatiotemporal predictive modeling. This paper proposes a geometry-aware active learning framework for modeling spatiotemporal dynamic systems. Specifically, we propose a geometry-aware spatiotemporal Gaussian Process (G-ST-GP) to effectively integrate the temporal correlations and geometric manifold features for reliable prediction of high-dimensional dynamic behaviors. In addition, we develop an adaptive active learning strategy to strategically identify informative spatial locations for data collection and further maximize the prediction accuracy. This strategy achieves the adaptive trade-off between the prediction uncertainty in the G-ST-GP model and the space-filling design guided by the geodesic distance across the 3D geometry. We implement the proposed framework to model the spatiotemporal electrodynamics in a 3D heart geometry. Numerical experiments show that our framework outperforms traditional methods lacking the mechanism of geometric information incorporation or effective data collection.
  \end{abstract}

  \noindent%

  \spacingset{1.5}

  % \section*{Note to Practitioners}
  % This article proposes an effective framework for predictive modeling of dynamic systems that involve complex 3D geometries. The key innovation of our framework lies in incorporating geometric awareness in spatiotemporal predictive modeling and further leveraging adaptive active learning to enable automatic identification of the most informative locations for sequential data collection. While our implementation focuses on cardiac modeling applications, this approach can be valuable in various industrial settings involving spatiotemporal predictive modeling, such as structural health monitoring, environmental sensing, and manufacturing process control. By addressing the challenges of geometric complexity and data efficiency, our framework has the potential to advance predictive modeling of complex dynamic systems and open new avenues for diverse practical applications.

  \section{Introduction} \label{s:intro}

Spatiotemporal dynamic systems are ubiquitous in science and engineering, governing phenomena such as cardiac electrical activities \cite{yang2023sensing, sim2020epicardial}, dispersion behaviors of pollutants \cite{liu2023inverse}, infectious disease spread \cite{velasquez2020forecast,liang2017climate}, and urban traffic flow \cite{wang2023latent}. These systems are featured with complicated dependencies between spatial distributions and temporal evolution, often exhibiting nonlinear and high-dimensional dynamics \cite{chen2015sparse}. Rapid advancements in sensing and imaging technologies provide unprecedented opportunities to collect signals across both the spatial and temporal domains to enhance the information visibility of the complex dynamic systems\cite{yao2021constrained,wang2021knowledge}. For example, cardiac catheterization procedures enable real-time monitoring of intracardiac electrical signals within the heart chamber, providing critical insights into arrhythmia and informing treatment strategies \cite{feltes2011indications}. Similarly, spatiotemporal transportation data can be collected from interconnected sensors, cameras, and GPS-enabled devices to track vehicle movements in real time over various street network layouts, offering opportunities to uncover traffic patterns for optimal routing and congestion management \cite{liu2017grid}. With the vast amount of spatiotemporal data, there is a pressing need to develop advanced analytical methods to fully realize the data potential for predictive modeling and control.

However, spatiotemporal modeling poses unique challenges due to the complex structure, high-dimensionality of sensor observations, nonlinear interactions across both the spatial and temporal domains, and inherent complexities in data collection such as sparse spatial sensor coverage. Specifically, realizing the full potential of spatiotemporal data for decision support calls upon coping with the following research challenges:

(1) High-dimensional sensing signals on complex geometries: Sensing and imaging of dynamic systems often generate high-dimensional signals collected from a vast number of time points and spatial locations distributed over complex geometries (e.g., multi-channel electrical signals on the heart surface \cite{chen2019characteristics,xie2022physics,yao2016physics}). Such high-dimensional data typically involves numerous variables and complex inter-dependencies, making it difficult to process and interpret. Additionally, the geometric complexity adds another layer of difficulty in predictive modeling of dynamic systems. For example, complex systems can involve irregular shapes, mesh data with unstructured grids, Riemannian manifolds, or non-Euclidean space (e.g., biological tissues) \cite{chen2016numerical,yao2016mesh,calandra2016manifold,xie2024kronecker,clayton2011models} that traditional modeling techniques struggle to accommodate.

(2) Expensive data collection process: Data collection in spatiotemporal systems is often costly and constrained by practical limitations \cite{calinon2007learning, yue2020active,xie2022physics2}. For example, during cardiac catheterization procedures, only a limited number of locations within the heart chamber can be queried for data collection due to procedural complexity, patient safety concerns, and time constraints \cite{vefali2020comparison,yao2024multi}. This underscores the importance of strategically selecting the most informative locations for data collection to maximize the value of the collected signals for improving the performance of predictive modeling. Such strategies are particularly vital in real-world applications, where balancing model accuracy with resource constraints is essential for both prediction reliability and practical feasibility.

This paper presents an effective geometry-aware active learning framework via Gaussian Process for predictive modeling of spatiotemporal dynamic systems. Our specific contributions are listed as follows:

\noindent(1) We develop a novel geometry-aware spatiotemporal Gaussian Process (G-ST-GP) framework for predictive modeling of dynamic systems. Our approach constructs the spatial kernel of the GP based on the eigenspace of the manifold Laplacian, leveraging the advances in reduced-rank GP modeling \cite{solin2020hilbert}, to effectively incorporate geometric information of 3D systems into the spatiotemporal modeling framework.

\noindent(2) We propose an effective adaptive active learning (A-AL) strategy to identify informative spatial locations for data acquisition, thereby maximizing the prediction capabilities of our G-ST-GP model. Our A-AL strategy incorporates both the prediction uncertainty in G-ST-GP and a space-filling design guided by the geodesic distance across the 3D geometry, the effects of which are adaptively balanced throughout the active learning process.

\noindent(3) We exploit the properties of Kronecker product of the spatiotemporal kernel in our G-ST-GP model to improve computational efficiency. By leveraging Kronecker operations, the posterior predictive modeling and adaptive active learning within the G-ST-GP framework can be achieved in a computation-efficient manner.

We validate our G-ST-GP framework by applying it to spatiotemporal predictive modeling of cardiac electrodynamics in a 3D ventricular geometry. Numerical experiments highlight the superior prediction performance of the proposed G-ST-GP model compared with traditional approaches that lack geometric information integration and efficient data collection capabilities.

The remainder of this paper is organized as follows: Section~\ref{s:re-review} presents the review of spatiotemporal predictive modeling in complex dynamic systems. Section~\ref{s:method} introduces the proposed G-ST-GP methodological framework. Section~\ref{s:results} demonstrates the effectiveness of our G-ST-GP method through numerical experiments in predictive modeling of cardiac electrodynamics, validating its performance for both prediction and active learning tasks. Finally, Section~\ref{s:conclusions} concludes the present investigation.

  \section{Research background} \label{s:re-review}

Spatiotemporal statistical methods have been vastly developed for the modeling and control of dynamic systems \cite{hamdi2022spatiotemporal}. Gaussian processes (GP) \cite{williams2006gaussian,liu2020gaussian} are extensively used in spatiotemporal modeling due to their flexibility in capturing spatial and temporal variations. Spatiotemporal GPs rely on covariance kernels that encode the dependency structure over space and time, enabling accurate space-time predictions. A fundamental component of kernel design is the definition of the distance metric used to measure the relationship between data points. The Euclidean distance remains the most commonly employed metric in this context and is widely adopted in kernel construction. For example, Koh \textit{et al.} developed a spatiotemporal log-Gaussian Cox process based on Euclidean distance in the spatial domain to model the wildfire size and occurrence intensity \cite{koh2023spatiotemporal}. Zhang \textit{et al.} developed a spatiotemporal GP for efficient prediction of weather data, whose spatial kernel is constructed based on Euclidean distance. Datta \textit{et al.} proposed a dynamic nearest-neighbor GP model based on Euclidean distance metric for both the spatial and temporal domains to facilitate the analysis of large space-time data for predicting particulate matter\cite{datta2016nonseparable}.

While the above applications of Euclidean metric-based GPs demonstrate promising outcomes, their reliance on Euclidean distance—which measures straight-line distances between points—limits their capability to account for detailed geometric information or model complex, unstructured spatial relations on manifolds, such as cardiac surfaces \cite{coveney2020gaussian,yao2024multi,yao2021spatiotemporal,yao2016mesh}. In such physical systems, meaningful distance measurements must consider manifold geometry and essential features, including surface distances and angles, or the frequencies and basis functions related to topological properties. For example, geodesic distance, defined as the shortest path on a 3D geometry connecting two spatial locations, is widely recognized to be effective in preserving geometric information. However, directly replacing Euclidean distances with geodesic distances will result in a non-positive semi-definite kernel matrix \cite{borovitskiy2020matern}, leading to the failure of GP modeling. Another approach is to project the dynamics on a 3D manifold into a 2D Euclidean embedding space by preserving the geodesic distance and then use the Euclidean distance-based kernel for predictive modeling \cite{xie2024kronecker}. However, geodesic distance alone is limited to fully characterize the geometric features such as frequencies and basis functions related to topological properties of the spatiotemporal dynamic systems.

Graph-based kernel design offers another approach to capture the local geometric structures in the spatial domain for spatiotemporal predictive modeling \cite{kriege2020survey}. For example, Wang \textit{et al.} developed a stochastic spatiotemporal model by incorporating the Kriging model into the Gaussian Markov random field (GMRF) to predict the dynamic thermal distribution in a 3D grid-like lattice \cite{wang2019modeling}. Hu \textit{et al.} constructed a spatiotemporal Graph Neural Process by modeling the spatial domain as a graph to predict the signals at target locations from neighboring contexts in the graph \cite{hu2023graph}. Javaheri proposed to integrate the GMRF with the temporal auto-regressive models to capture the spatiotemporal dependencies for graph learning \cite{javaheri2024learning}. Note that graph-based spatiotemporal GP modeling designs the spatial kernel using local neighborhood structures to capture correlations between adjacent sites. However, by assuming zero correlation between distant observations, those models focus solely on local geometric features while losing global information on 3D geometries or manifolds.

In addition to statistical approaches, advanced deep learning architectures \cite{wang2020deep,wang2021multi,wang2023hierarchical} have been designed to investigate spatiotemporal data. For example, Convolutional Neural Networks (CNNs) are designed to capture spatial features, while recurrent neural networks (RNNs) and temporal attention mechanisms are used to account for temporal dynamics \cite{wang2020deep}. Hybrid models combining these architectures such as ConvLSTM \cite{azad2019bi} have demonstrated strong performance in applications including spatiotemporal weather forecasting \cite{wang2019deep} and dynamics prediction \cite{zhao2022deep}. Moreover, spatiotemporal graph neural network models \cite{yu2017spatio} have been designed to address the limitations of traditional ConvLSTMs that are only applicable to model grid-like data such as time-varying imaging data. For example, dynamic GNNs have been developed for spatiotemporal predictive modeling in non-Euclidean space for traffic flow forecasting \cite{ali2022exploiting}. 
 
However, one significant limitation of spatiotemporal deep learning models such as ConvLSTMs or spatiotemporal GNNs is their inherent dependency on spatial structures or kernels that are typically constructed around measurement sites, limiting their predictive capability to these predefined nodes. As such, predictions are not readily extendable to arbitrary spatial locations where data might be sparse or unavailable. This limitation poses challenges in applications requiring fine-grained spatial resolution or interpolations in regions outside the immediate vicinity of the measurement points. Furthermore, in addition to their heavy memory and inference cost, deep learning models are widely recognized to be limited in effective uncertainty quantification \cite{liu2020simple}, which poses significant challenges in designing robust active learning strategies to increase the predictive power of the model in situations when data acquisition is expensive. Thus, there is an urgent need to develop an effective spatiotemporal predictive modeling approach by capturing the intrinsic geometric features of the spatial domain and the dynamic nature of the temporal evolution, which can further be leveraged to design robust and adaptive active learning strategies for maximizing predictive accuracy while minimizing data acquisition costs in resource-intensive scenarios.

  \section{Research Methodology} \label{s:method}

\begin{figure*}
	\centering
	\includegraphics[width=6in]{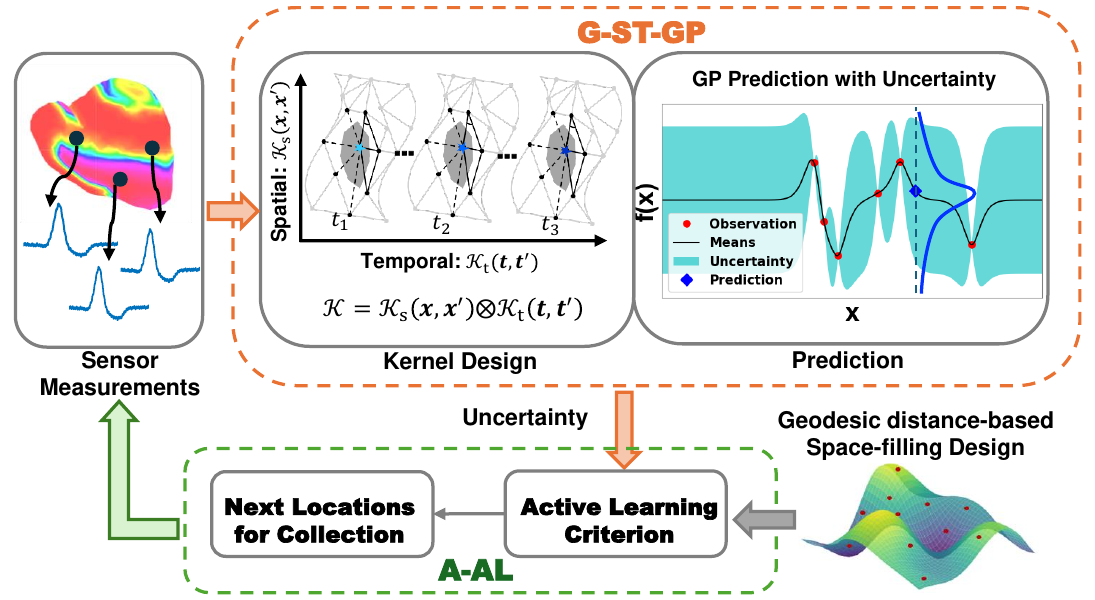}
    \caption{Flowchart of the proposed methodology. The G-ST-GP framework is first developed to model the spatiotemporal dynamics in a 3D complicated geometry. Then, an adaptive active learning method (A-AL) is designed by combining the prediction uncertainty in G-ST-GP with a space-filling design guided by geodesic distance to identify informative locations for additional signal measurements, enhancing the predictive capability of our G-ST-GP model.}
	\label{fig:flowchart}
\end{figure*} 

This section presents the proposed geometry-aware active learning framework as illustrated in Fig.~\ref{fig:flowchart}. The framework consists of two main components. First, we develop a geometry-aware spatiotemporal Gaussian Process (G-ST-GP) to effectively incorporate the geometric information and temporal evolution to model complex dynamic systems. Second, we introduce an effective active learning approach that integrates the prediction uncertainty in G-ST-GP with space-filling design guided by the geodesic distance across the 3D geometry to identify informative spatial locations for subsequent signal measurements, thereby enhancing the predictive capabilities of our G-ST-GP model.
    
A GP is a stochastic process with an infinite number of random variables such that every finite subset of the variables follows a joint multivariate Gaussian distribution. In our study, we denote the variable describing the underlying system dynamics at location $\vect{x}$ and time $t$ as $u(\vect{x},t)$, with the corresponding measurement denoted as $y(\vect{x},t)$. Our proposed G-ST-GP models $u(\vect{x},t)$ and $y(\vect{x},t)$ as:
    \begin{eqnarray}
     y(\vect{x},t) &=& u(\vect{x},t)+\epsilon(\vect{x},t) \\
     u &\sim& \mathcal{GP}(0,\mathcal{K}_\text{st}) \\
     \epsilon &\sim& \mathcal{GP}(0,\mathcal{K}_{\epsilon,\text{st}})
    \end{eqnarray}
    where $\epsilon(\vect{x},t)$ is a spatiotemporal noise term or nugget effect. In order to capture the spatiotemporal interactions, we propose to design the kernel function for $u(\vect{x},t)$ as $\mathcal{K}_\text{st} = \mathcal{K}_\text{s} \otimes \mathcal{K}_\text{t}$, where $\mathcal{K}_\text{s}$ denotes the spatial kernel to capture the spatial covariances in a 3D complicated geometry and $\mathcal{K}_\text{t}$ is the temporal kernel to capture covariances in the time domain. Additionally, $\mathcal{K}_{\epsilon,\text{st}}$ describes the spatiotemporal covariance in $\epsilon(\vect{x},t)$, which is assumed to have the form: $\mathcal{K}_{\epsilon,\text{st}}(\vect{x}_1,\vect{x}_2; t_1,t_2)=\sigma_{\epsilon,\text{s}}\mathcal{K}_\text{t}(t_1,t_2)+\sigma_{\epsilon,t}\mathcal{K}_\text{s}(\vect{x}_1,\vect{x}_2)+\sigma_{\epsilon,\text{s}}\sigma_{\epsilon,t}$.
 
Given the observed training data, $\vect{y}_\text{tr}=[y(\vect{x}_i,t_j)]_{\vect{x}_i\in\mathcal{X}_\text{tr},t_j\in\mathcal{T}_\text{tr}}\subset \mathbb{R}^{N_\text{s}N_\text{t}}$, collected at the spatial locations $\mathcal{X}_\text{tr}=\{\vect{x}_1,\dots,\vect{x}_{N_\text{s}}\}$ and time points $\mathcal{T}_\text{tr}=\{t_1,\dots,t_{N_\text{t}}\}$, the marginal distribution of $\vect{y}_\text{tr}$ is given by:
    \begin{eqnarray}
        \vect{y}_\text{tr}\sim\mathcal{N}(\vect{0},\Sigma_\text{tr})
        \label{Eq: marginal}
    \end{eqnarray}
    with $\Sigma_\text{tr}=(\mathcal{K}_\text{s}(\mathcal{X}_\text{tr},\mathcal{X}_\text{tr})+\sigma_{\epsilon,\text{s}}\mathcal{I}_{N_\text{s}})\otimes(\mathcal{K}_\text{t}(\mathcal{T}_\text{tr},\mathcal{T}_\text{tr})+\sigma_{\epsilon,t}\mathcal{I}_{N_\text{t}})$ denoting the covariance of $\vect{y}_\text{tr}$, where $\mathcal{I}_{N_\text{s}}$ ($\mathcal{I}_{N_\text{t}}$) is an $N_\text{s}\times N_\text{s}$ ($N_\text{t}\times N_\text{t}$) identity matrix. Given collected observations $[\vect{y}_\text{tr};\mathcal{X}_\text{tr},\mathcal{T}_\text{tr}]$, the predictive distribution of the dynamics at an arbitrary spatiotemporal coordinate $(\vect{x}^*,t^*)$ is:
    \begin{equation}
    \begin{aligned}
        u(\vect{x}^*,t^*) | \vect{y}_\text{tr};\mathcal{X}_\text{tr},\mathcal{T}_\text{tr} &\sim 
        \mathcal{N} \Big( \mathcal{K}_\text{st}^* \Sigma_\text{tr}^{-1} \vect{y}_\text{tr}, \mathcal{K}_\text{st}^{**} -  \mathcal{K}_\text{st}^* \Sigma_\text{tr}^{-1} ( \mathcal{K}_\text{st}^* )^\top \Big)
    \end{aligned}
    \label{eq:trivial-GP}
    \end{equation}
    where $ \mathcal{K}_\text{st}^* =\mathcal{K}_\text{st}\left((\vect{x}^*,t^*);(\mathcal{X}_\text{tr},\mathcal{T}_\text{tr})\right)$ and $ \mathcal{K}_\text{st}^{**} =\mathcal{K}_\text{st}\left((\vect{x}^*,t^*);(\vect{x}^*,t^*)\right)$.  As such, the model's predictive performance fundamentally depends on the construction of effective kernels, $\mathcal{K}_\text{s}$ and $\mathcal{K}_\text{t}$, which will be detailed in the following sub-sections.

\subsection{\emph{Spatial Kernel modeling}} \label{s:methods.1}
The spatial kernel $\mathcal{K}_\text{s}$ should be defined in a way to capture the geometric characteristics of the 3D systems to effectively describe the spatial correlations. We propose to leverage the advances in reduced-rank GP modeling \cite{solin2020hilbert} to define the spatial kernel matrix in terms of eigen-functions of the Laplacian operator on the 3D geometry. Let $\Omega \subset \mathbb{R}^d$ be a compact set, where $d$ is the dimension of the space. For all $\vect{x} \in {\Omega}$, we define a general covariance operator $\mathcal{K}$ as \cite{solin2020hilbert}:
    \begin{equation}
        \begin{aligned}
            \mathcal{K} \, f(\vect{x}) := \int_{\Omega} \mathcal{K}(\vect{x}, \vect{x}') \, f(\vect{x}') \, \mathrm{d} \vect{x}' \label{eq:def-K}
        \end{aligned}
    \end{equation}
    where $\mathcal{K}(\vect{x}, \vect{x}')$ denotes the covariance function (kernel) with inputs $\vect{x}$ and $\vect{x}'$, and $f(\cdot)$ is a function defined on $\Omega$. %For the stationary case, the covariance operator $\mathcal{K}$ can be formulated as the Fourier transform of a positive finite measure on $\Omega$ \cite{williams2006gaussian}. Additionally,  $\mathcal{K}$ admits a basis decomposition in Hilbert space. We will combine the Fourier transform and basis representations to construct a spatial kernel to capture the geometric characteristics of the 3D systems, which will be elaborated in the subsequent sub-sections.

\subsubsection{\emph{Series Expansion of the Spatial Covariance Operator}}
In the stationary case, the covariance function only relies on the difference between two input vectors, i.e., $\mathcal{K}(\vect{x}, \vect{x}')=\mathcal{K}(\vect{r})$ with $\vect{r} = \vect{x} - \vect{x}'$, exhibiting translation invariance. According to Bochner's theorem \cite{stein1999interpolation}, $\mathcal{K}(\vect{r})$ can be represented as:
    \begin{equation}
        \begin{aligned}
            \mathcal{K}(\vect{r}) = \int_{\Omega} e^{2 \pi i \vect{s}^{\intercal} \vect{r}} \, \mathrm{d}\mu(\vect{s})
        \end{aligned}
        \label{Eq: k(r)}
    \end{equation}
    where $\vect{s}$ represents the frequency vector in the Fourier domain and $\mu$ is a positive finite measure on $\Omega$. Furthermore, given the density $S(\vect{s})$ of measure $\mu$, which is also known as the spectral density corresponding to $\mathcal{K}(\vect{r})$, the Wiener-Khintchine theorem \cite{chatfield1989wktheory} provides an alternative formulation of Eq. (\ref{Eq: k(r)}):
    \begin{equation}
        \begin{aligned}
            \mathcal{K}(\vect{r}) &= \int S(\vect{s}) e^{2 \pi i \vect{s}^{\intercal} \vect{r}} \, \mathrm{d}\vect{s}
        \end{aligned}
        \label{Eq: k(s)}
    \end{equation}
    with its inverse Fourier relationship:
    \begin{equation}
        \begin{aligned}
            S(\vect{s}) &= \int \mathcal{K}(\vect{r}) e^{-2 \pi i \vect{s}^{\intercal} \vect{r}} \, \mathrm{d}\vect{r}
        \end{aligned}
        \label{Eq: S}
    \end{equation}

Eqs. (\ref{Eq: k(s)})-(\ref{Eq: S}) establish the Fourier duality between the covariance function $\mathcal{K}(\vect{r})$ and the spectral density $S(\vect{s})$. Notably, the spectral density of the Gaussian process $S(\vect{s})$ depends solely on the Euclidean norm of the frequency vector $\|\vect{s}\|$ \cite{solin2020hilbert}. For an analytical $S(\vect{s})$, we can express it as a polynomial series:
    \begin{equation}
        \begin{aligned}
            S(\| \vect{s} \|) = \sum_{p \in \mathbb{N}} a_p (\| \vect{s} \|^2)^p
        \end{aligned}
        \label{eq:S-span}
    \end{equation}
    where $a_p$ are the series coefficients and $\mathbb{N}$ denotes the set of natural numbers. Moreover, the Laplacian operator $\Delta$ acting on a function $f(\vect{x})$ ($\vect{x}\in \Omega$) is given as:
    \begin{equation}
        \begin{aligned}
             \Delta f(\vect{x}) = \sum_{i=1}^d \frac{\partial^2 f(\vect{x})}{\partial x_i^2}
        \end{aligned}
    \end{equation}
    Applying the Fourier transform to both sides yields:
    \begin{equation}
        \begin{aligned}
            \mathcal{F}[\Delta f](\vect{s}) &= \sum_{i=1}^d \mathcal{F}\left[\frac{\partial^2 f}{\partial x_i^2}\right](\vect{s})
            = - \sum_{i=1}^d ( s_{i})^2 \mathcal{F}[f](\vect{s}) = -\|\vect{s}\|^2 \mathcal{F}[f](\vect{s})
        \end{aligned}
        \label{eq:trans-ftn}
    \end{equation}
    where $\mathcal{F}[f](\vect{s})$ denotes the Fourier transform of $f$. Eq.\eqref{eq:trans-ftn} demonstrates that the transfer function of the Laplacian operator $\Delta$ is $-\|\vect{s}\|^2$. Substituting Eq.\eqref{eq:trans-ftn} into Eq.\eqref{eq:S-span}, and taking the inverse Fourier transform, we obtain:
    \begin{equation}
        \begin{aligned}
            \mathcal{K}= \mathcal{F}^{-1} \left[ S(\| \vect{s} \|) \right] = \mathcal{F}^{-1} \left[ \sum_{p \in \mathbb{N}} a_p (\| \vect{s} \|^2)^p \right]= \sum_{p \in \mathbb{N}} a_p (-\Delta)^p
        \end{aligned}
        \label{eq:K-expan-Fou}
    \end{equation}
    which is the series expansion of the covariance operator $\mathcal{K}$ in terms of powers of the Laplacian operator $\Delta$.

\subsubsection{\emph{Decomposition of Covariance Operator under Eigen-basis}}
The covariance operator $\mathcal{K}$ can be further approximated by a basis decomposition in Hilbert space \cite{solin2020hilbert}. Specifically, the eigenproblem for the Laplacian operator (which is a positive definite Hermitian operator) with Neumann boundary conditions is:
    \begin{eqnarray}
       -\Delta \phi_j(\vect{x}) &=& \lambda_j \phi_j(\vect{x}), \quad \vect{x} \in \Omega, \quad j \in \mathbb{N}^* \nonumber\\
       \vect{n}_{\partial\Omega}\cdot \nabla \phi_j(\vect{x}) &=& 0, \quad \vect{x} \in \partial\Omega, \quad j \in \mathbb{N}^*
       \label{eq:eigen-solver} 
    \end{eqnarray}
    where $\lambda_j$ and $\phi_j(\vect{x})$ are the eigenvalues and eigenfunctions respectively, $\mathbb{N}^*$ is the set of positive natural numbers, $\partial\Omega$ denotes the boundary of domain $\Omega$, and $\vect{n}_{\partial\Omega}$ is the normal vector on $\partial\Omega$. Note that the Neumann boundary condition is selected to impose zero normal derivative at the boundary of the domain. This type of boundary condition is commonly used in modeling reaction-diffusion dynamics, particularly when there is no external flow or gradient at the boundary, e.g., cardiac electrodynamics within the heart chamber \cite{yang2023sensing}. 

Eigenfunctions $\phi_j(\vect{x})$ possess the orthonormality property:
    \begin{equation}
        \begin{aligned}
           \langle \phi_i(\vect{x}), \phi_j(\vect{x}) \rangle=\int_{\Omega}\phi_i(\vect{x})\phi_j(\vect{x})\mathrm{d}\vect{x} = \delta_{ij}, \quad i,j \in \mathbb{N}^*
        \end{aligned}
    \end{equation}
    where $\delta_{ij}$ is the Kronecker delta, which equals to $1$ if and only if $i=j$ and $0$ otherwise, enabling the eigenvectors to serve as an orthogonal basis in Hilbert space. Given the eigenfunctions of $\Delta$, an arbitrary function $f(\vect{x})$ can be written as a linear combination of those eigenfunctions, i.e., $f(\vect{x})=\sum_{j \in \mathbb{N}^*}\langle \phi_j(\vect{x}),f(\vect{x})\rangle \phi_j(\vect{x})$. Hence, the negative Laplacian operator satisfies:
    \begin{equation}
        \begin{aligned}
            -\Delta f(\vect{x}) = \int_{\Omega} l(\vect{x}, \vect{x}') \, f(\vect{x}') \, \mathrm{d}\vect{x}' \label{eq:Lap-1dim}
        \end{aligned}
    \end{equation}
    with
    \begin{equation}
        \begin{aligned}
            l(\vect{x}, \vect{x}') = \sum_{j \in \mathbb{N}^*} \lambda_j \, \phi_j(\vect{x}) \, \phi_j(\vect{x}') \label{eq:l-1dim}
        \end{aligned}
    \end{equation}
    where $l(\vect{x}, \vect{x}')$ can be viewed as the kernel function associated with the Laplacian operator. Note that Eqs.~\eqref{eq:Lap-1dim} and \eqref{eq:l-1dim} can be generalized to higher powers of the Laplacian operator:
    \begin{equation}
        \begin{aligned}
            \left( -\Delta \right)^p f(\vect{x}) = \int_{\Omega} l^p(\vect{x}, \vect{x}') \, f(\vect{x}') \, \mathrm{d}\vect{x}',  \quad p \in \mathbb{N}
        \end{aligned} \label{eq:Lap-ndim}
    \end{equation}
    \begin{equation}
        \begin{aligned}
            l^p(\vect{x}, \vect{x}') = \sum_{j \in \mathbb{N}^*} (\lambda_j)^p \, \phi_j(\vect{x}) \, \phi_j(\vect{x}'),  \quad p \in \mathbb{N} \label{eq:l-ndim}
        \end{aligned}
    \end{equation}
    Substituting Eq.(\ref{eq:Lap-ndim}) and Eq.(\ref{eq:l-ndim}) into the previous expansion in Eq. (\ref{eq:K-expan-Fou}) yields:
    \begin{equation}
        \begin{aligned}
            \mathcal{K} f(\vect{x}) &= \sum_{p \in \mathbb{N}} a_p \left( -\Delta \right)^p f(\vect{x}) = \int_{\Omega} \sum_{p \in \mathbb{N}} a_p \, l^p(\vect{x}, \vect{x}') f(\vect{x}') \, \mathrm{d}\vect{x}' \\
            &= \int_{\Omega} \sum_{p \in \mathbb{N}} \sum_{j \in \mathbb{N}^*} a_p (\lambda_j)^p \, \phi_j(\vect{x}) \, \phi_j(\vect{x}') \, f(\vect{x}') \, \mathrm{d}\vect{x}' \label{eq:K-expan-Eigen}
        \end{aligned}
    \end{equation}

Comparing the initial definition of the covariance operator $\mathcal{K}$ given by Eq. (\ref{eq:def-K}) with Eq.(\ref{eq:K-expan-Eigen}), we can express the covariance function $\mathcal{K}(\vect{x}, \vect{x}')$ as:
    \begin{equation}
        \begin{aligned}
            \mathcal{K}(\vect{x}, \vect{x}') = \sum_{p \in \mathbb{N}} \sum_{j \in \mathbb{N}^*} a_p (\lambda_j)^p \, \phi_j(\vect{x}) \, \phi_j(\vect{x}')
        \end{aligned}
    \end{equation}
    According to the definition of $S(\|\vect{s}\|)$ in Eq. (\ref{eq:S-span}) and letting $\|\vect{s}\|=\sqrt{\lambda_j}$, we can approximate $\mathcal{K}(\vect{x}, \vect{x}')$ as:
    \begin{equation}
        \begin{aligned}
            \mathcal{K}(\vect{x}, \vect{x}') \approx \sum_{j \in \mathbb{N}^*} S\left( \sqrt{\lambda_j} \right) \, \phi_j(\vect{x}) \, \phi_j(\vect{x}') \label{eq:sp-def}
        \end{aligned}
    \end{equation}
    where $S(\sqrt{\lambda_j})$ is the spectral density evaluated at the square root of the eigenvalues $\lambda_j$. According to Eq. (\ref{eq:sp-def}), after specifying the spectral density $S(\cdot)$, the kernel function $\mathcal{K}(\vect{x}, \vect{x}')$ is fully characterized by the Laplacian operation $\Delta$. Thus, incorporating geometry information into the kernel function calls upon the careful derivation of $\Delta$ by accounting for the geometric characteristics of the complex system.

\subsubsection{\emph{Spatial Kernel Construction via the Laplacian Analysis on the 3D Geometry}}

\begin{figure}
\centering
\includegraphics[width=0.5\linewidth]{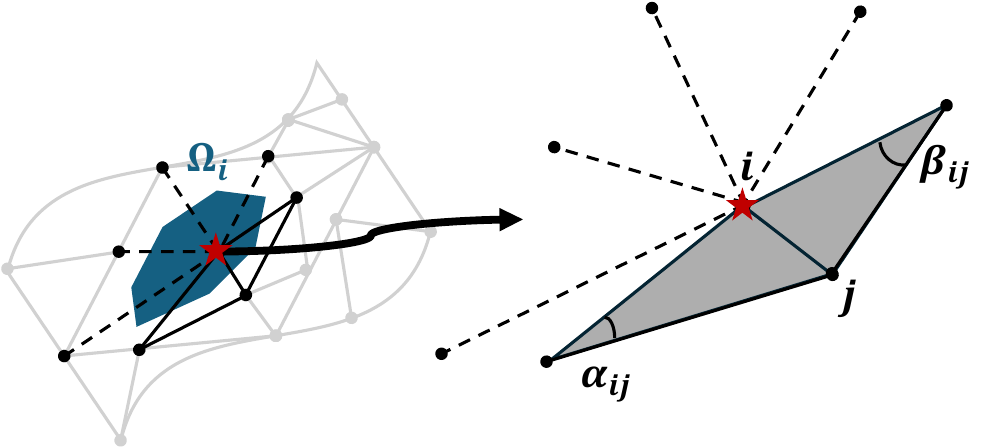}
\caption{Illustration of angles $\alpha_{ij}$ and $\beta_{ij}$ associated with the edge connecting vertices $i$ and $j$, and area $\Omega_i$ corresponding to vertex $i$ in Eq.(\ref{eq:angle}).}
\label{fig:LapMat}
\end{figure}

We propose to leverage the Laplacian operator of the 3D geometry to capture the geometric features, such as connectivity between discretized vertices and angles between edges, and further facilitate effective spatial kernel design using the eigen-decomposition analysis as illustrated in Eq.(\ref{eq:sp-def}). Specifically, considering a discretized manifold geometry $M$ in 3D space with $N$ vertices, the Laplacian operator $\Delta_M$ on the dynamics $u$ at vertex $i$ is given as \cite{sorkine2005laplacian}:
    \begin{eqnarray}
        (\Delta_M u)_i=\frac{1}{2|\Omega_i|}\sum_{j \in N(i)}(\cot \alpha_{ij} + \cot \beta_{ij})(u_j-u_i)
        \label{eq:angle}
    \end{eqnarray}
    where $i, j = 1, \ldots, N$ index the vertices, $|\Omega_i|$ is the area of Voronoi cell corresponding to vertex $i$, $N(i)$ denotes the set of neighboring vertices connected to vertex $i$, and $\alpha_{ij}$ and $\beta_{ij}$ are the angles opposite the edge between vertices $i$ and $j$ (see Fig.~\ref{fig:LapMat}). Considering the Neumann boundary condition given in Eq. (\ref{eq:eigen-solver}), the eigenvalue problem for the original Laplacian operator becomes the eigenvalue problem for the Manifold Laplacian (the Laplacian restricted to the manifold, denoted as $\Delta_M$), i.e., $-\Delta_M\phi =\lambda\phi$. This is true because 
    \begin{equation}
    \begin{aligned}
          \Delta_M \phi&=\nabla_M\cdot \nabla_M\phi=\nabla_M\cdot\left[\nabla\phi-(\vect{n}_M\cdot \nabla \phi)\vect{n}_{M}\right]
        =\nabla_M\cdot\nabla\phi=\nabla\cdot\nabla\phi=\Delta\phi
    \end{aligned}
    \end{equation}
    where $\nabla_M$ denotes the gradient operator on manifold $M$, $\vect{n}_{M}$ is the normal vector on $M$, the third equality is valid because of the Neumann boundary condition, and the fourth equality is true because $\nabla\phi$ is tangent to $M$ if the boundary condition is satisfied. 

Using the eigenfunctions $\phi(\cdot)$ and eigenvalues $\lambda$ of $\Delta_M$, we can formally define the spatial kernel corresponding to the spatial covariance operator $\mathcal{K}_{\text{s}}$ with a scaling coefficient $\sigma_{m}$:
    \begin{equation}
        \mathcal{K}_{\text{s}}(\vect{x}, \vect{x}') = \sum_{j \in \mathbb{N}^*} \sigma_{m} S\left( \sqrt{\lambda_j} \right) \, \phi_j(\vect{x}) \, \phi_j(\vect{x}')  
        \label{eq: def-kernel-spa}
    \end{equation}
    where $S(\sqrt{\lambda})$ is a non-increasing function of $\lambda$, and $\lambda_1\leq \lambda_2\leq\dots$.
    In this study, we adopt the Matérn covariance structure for predictive modeling, whose spectral density function $S(\sqrt{\lambda})$ is:
    \begin{equation}
        % \begin{aligned}
            S(\sqrt{\lambda}) := \frac{2^d \pi^{d/2} \Gamma(\nu + d/2) (2 \nu)^{\nu}}{\Gamma(\nu) l_{\text{s}}^{2 \nu}} \left( \frac{2 \nu}{l_{\text{s}}^2} + 4 \pi^2 \lambda \right)^{-(\nu + d/2)}
        % \end{aligned}
    \end{equation}
    where $d$ is the dimensionality (which is 2 for the manifold or meshed geometry in 3D space), $\Gamma$ represents the Gamma function, $\nu$ controls the smoothness (which is selected as $3/2$ in our study), and $l_{\text{s}}$ is the spatial length-scale parameter.

\subsection{\emph{Temporal Kernel Construction}}

    In addition to the spatial distribution, the dynamic process is also evolving over time with temporal correlation. In this work, we employ the Matérn kernel with smoothness parameter $\nu=3/2$ to capture the temporal correlation and further facilitate spatiotemporal predictive modeling. Specifically, the temporal kernel is defined as follows:
    \begin{equation}
        \begin{aligned}
            \mathcal{K}_{\text{t}}(t, t') &= \sigma_a \left( 1 + \frac{\sqrt{3}\|t - t'\|}{l_{\text{t}}} \right) \exp\left( - \frac{\sqrt{3}\|t - t'\|}{l_{\text{t}}} \right)  \label{eq:def-kernel-tem}
        \end{aligned}
    \end{equation}
    where $l_\text{t}$ denotes the length scale and $\sigma_a$ is the scaling coefficient for the temporal kernel.

\subsection{\emph{G-ST-GP Posterior Modeling}}

\subsubsection{\emph{G-ST-GP Parameter Estimation}}
We denote the G-ST-GP parameter set as $\vect{\theta}=\{l_{\text{s}}, \sigma_m, \sigma_{\epsilon,\text{s}}, l_{\text{t}}, \sigma_a, \sigma_{\epsilon,\text{t}}\}$, which will be estimated using the maximum likelihood method. Given the collected training data $[\vect{y}_\text{tr};\mathcal{X}_\text{tr},\mathcal{T}_\text{tr}]$ and marginal distribution in Eq. (\ref{Eq: marginal}), the negative log marginal likelihood is:
    \begin{equation}
        \begin{aligned}
            -\log p(\vect{y}_\text{tr} | \mathcal{X}_\text{tr},\mathcal{T}_\text{tr};\mathbf{\theta}) 
            &= \frac{1}{2} \vect{y}_\text{tr}^\top \Sigma_\text{tr}^{-1} \vect{y}_\text{tr} + \frac{1}{2} \log \left|\Sigma_\text{tr} \right| + \frac{N_\text{s}N_\text{t}}{2} \log 2 \pi
            \label{eq:NLL}
        \end{aligned}
    \end{equation}
    where $N_\text{s}N_\text{t}$ is the total number of spatiotemporal observations. Eq.~\eqref{eq:NLL} comprises three terms: a data fit term $f_1$, a complexity penalty $f_2$, and a normalization constant $f_3$. Note that considering a spatial kernel matrix $\mathcal{K}_{\text{s}}$ of dimension $(N_\text{s} \times N_\text{s})$ and a temporal kernel matrix $\mathcal{K}_{\text{t}}$ of dimension $(N_\text{t} \times N_\text{t})$, the resulting coupling matrices $\mathcal{K}_\text{st}$ and $\Sigma_\text{tr}$ are with the dimensions of $(N_\text{s}N_\text{t}) \times (N_\text{s}N_\text{t})$, introducing significant computational challenges in calculating the matrix inversion and determinant in $f_1$ and $f_2$. Here, we propose to improve the computational efficiency by exploiting the properties of Kronecker product:
    \begin{equation}
        \begin{aligned}
            f_1 :&= \frac{1}{2} \vect{y}_\text{tr}^\top \Sigma_\text{tr}^{-1} \vect{y}_\text{tr}=\frac{1}{2} \vect{y}_\text{tr}^\top \left( \Sigma_{\text{tr},\text{s}}\otimes \Sigma_{\text{tr},\text{t}} \right)^{-1} \vect{y}_\text{tr} 
            =\frac{1}{2} \vect{y}_\text{tr}^\top \left( \Sigma_{\text{tr},\text{s}}^{-1}\otimes \Sigma_{\text{tr},\text{t}}^{-1} \right) \vect{y}_\text{tr}  \\
            &= \frac{1}{2} \vect{y}_\text{tr}^\top \operatorname{vec} \left( \Sigma_{\text{tr},\text{t}}^{-1} \operatorname{vec}^{-1}(\vect{y}_\text{tr}) \Sigma_{\text{tr},\text{s}}^{-1} \right)\\
            f_2 :&= \frac{1}{2} \operatorname{log} \left|\Sigma_\text{tr} \right| = \frac{1}{2} \operatorname{log} \left| \Sigma_{\text{tr},\text{s}}\otimes \Sigma_{\text{tr},\text{t}} \right|
            = \frac{1}{2} \operatorname{log} \left( \left| \Sigma_{\text{tr},\text{s}} \right|^{N_\text{t}} \cdot \left| \Sigma_{\text{tr},\text{t}} \right|^{N_\text{s}} \right) \\
            &= \frac{1}{2} \left( N_\text{t}\operatorname{log} \left| \Sigma_{\text{tr},\text{s}} \right| + N_\text{s} \operatorname{log} \left| \Sigma_{\text{tr},\text{t}} \right| \right) 
    \end{aligned}
    \end{equation}
    where $\Sigma_{\text{tr},\text{s}}=\mathcal{K}_\text{s}(\mathcal{X}_\text{tr},\mathcal{X}_\text{tr})+\sigma_{\epsilon,\text{s}}\mathcal{I}_{N_\text{s}}$, $\Sigma_{\text{tr},\text{t}}=\mathcal{K}_\text{t}(\mathcal{T}_\text{tr},\mathcal{T}_\text{tr})+\sigma_{\epsilon,\text{t}}\mathcal{I}_{N_\text{t}}$, $\operatorname{vec}(\cdot)$ is the vectorization operator that stacks matrix columns into a single column vector, and $\operatorname{vec}(\cdot)^{-1}$ is the inverse operator of $\operatorname{vec}(\cdot)$. As such, minimizing Eq. (\ref{eq:NLL}) to estimate $\vect{\theta}$ will only involve the inverse and determinant computations of matrices with dimensions of $N_\text{s}\times N_\text{s}$ and $N_\text{t}\times N_\text{t}$, leading to a significant increase in computational efficiency.

\subsubsection{\emph{Posterior Prediction}}
Given the estimated parameters $\hat{\vect{\theta}}$ from the maximum likelihood method and by further exploiting the Kronecker properties, the posterior mean vector of the dynamics at a test set, $[(\vect{x}^*_i,t^*_j)]_{\vect{x}^*_i\in\mathcal{X}^*,t^*_j\in\mathcal{T}^*}$, is given as:
    \begin{equation}
        \begin{aligned}
            \hat{\vect{\mu}}^*(\mathcal{X}^*,\mathcal{T}^*) &= \left( \mathcal{K}_\text{s}^* \otimes \mathcal{K}_\text{t}^* \right) \left( \Sigma_{\text{tr},\text{s}}\otimes \Sigma_{\text{tr},\text{t}} \right)^{-1} \vect{y}_\text{tr}
            =\left( \mathcal{K}_\text{s}^* \Sigma_{\text{tr},\text{s}}^{-1} \right) \otimes \left( \mathcal{K}_\text{t}^* \Sigma_{\text{tr},\text{t}}^{-1} \right) \vect{y}_\text{tr}  \\
            &=  \operatorname{vec} \left[ \left( \mathcal{K}_\text{t}^* \Sigma_{\text{tr},\text{t}}^{-1} \right) \operatorname{vec}^{-1}(\vect{y}_\text{tr}) \left( \mathcal{K}_\text{s}^* \Sigma_{\text{tr},\text{s}}^{-1} \right)^T \right]
        \end{aligned}
        \label{Eq: mu}
    \end{equation}
    where $\mathcal{K}^*$ represents kernel matrix for test-training pairs. Additionally, the predictive variance vector for uncertainty quantification can be expressed as:
    \begin{equation}
        \begin{aligned}
            (\hat{\vect{\nu}}^{*})^{2}(\mathcal{X}^*,\mathcal{T}^*)
            &=  \operatorname{diag}\left(\left( \mathcal{K}_{\text{s}}^{**} \otimes \mathcal{K}_{\text{t}}^{**} \right) - \left( \mathcal{K}_{\text{s}}^* \otimes \mathcal{K}_{\text{t}}^* \right) \left( \Sigma_{\text{tr},\text{s}}\otimes \Sigma_{tr,t} \right)^{-1} \left( \mathcal{K}_{\text{s}}^* \otimes \mathcal{K}_{\text{t}}^* \right)^\mathrm{T}\right) \\
            &= \operatorname{diag}\left(\mathcal{K}_{\text{s}}^{**} \otimes \mathcal{K}_{\text{t}}^{**} - \left( \mathcal{K}_{\text{s}}^* \Sigma_{\text{tr},\text{s}}^{-1} (\mathcal{K}_{\text{s}}^*)^\mathrm{T} \right) \otimes \left( \mathcal{K}_{\text{t}}^* \Sigma_\text{tr,t}^{-1} (\mathcal{K}_{\text{t}}^*)^\mathrm{T} \right)\right)\\
            &= \operatorname{diag}(\mathcal{K}_{\text{s}}^{**}) \otimes \operatorname{diag}(\mathcal{K}_{\text{t}}^{**})
            - \operatorname{diag}\left( \mathcal{K}_{\text{s}}^* \Sigma_{\text{tr},\text{s}}^{-1} (\mathcal{K}_{\text{s}}^*)^\mathrm{T} \right) \otimes \operatorname{diag}\left( \mathcal{K}_{\text{t}}^* \Sigma_\text{tr,t}^{-1} (\mathcal{K}_{\text{t}}^*)^\mathrm{T} \right)
        \label{Eq: var}
    \end{aligned}
    \end{equation}
    where $\mathcal{K}^{**}$ represents kernels for test-test pairs and $\operatorname{diag}(\cdot)$ extracts the diagonal elements of a matrix.

\subsection{\emph{Adaptive Active Learning in G-ST-GP Modeling}}
The predictive capability of G-ST-GP also highly relies on the quality and volume of training data. In many real-world applications, particularly in medical domains, data collection can be costly and constrained by practical limitations. To achieve reliable predictive modeling of spatiotemporal systems under such conditions, it is essential to design an effective active learning strategy for optimal data acquisition by identifying the most informative data points, thereby enhancing model accuracy while minimizing data collection costs.

A widely used active learning strategy is the uncertainty-based sampling: the next spatial observation point is selected as the one with the highest predictive variance \cite{srinivas2009gaussian}:
    \begin{eqnarray}        \vect{x}_{n+1}=\arg\max_{\vect{x}}\hat{\nu}_n(\vect{x})
        \label{Eq: S-AL}
    \end{eqnarray}
    where $\hat{\nu}_n(\vect{x})$ represents the prediction uncertainty at unmeasured spatial location $\vect{x}$ given the measured signals collected from existing $n$ spatial locations. In our study, we estimate the prediction uncertainty at location $\vect{x}$ as
    \begin{eqnarray}  \hat{\nu}_n(\vect{x})=\frac{\sum_{j=1}^{N_\text{t}}\hat{\nu}_n(\vect{x},t_j)}{N_\text{t}}
    \end{eqnarray}
    where $\hat{\nu}_n(\vect{x},t_j)$ is the predictive standard deviation at  $(\vect{x},t_j)$ (see Eq. (\ref{Eq: var})) given by the G-ST-GP trained using dynamic signals collected from $n$ spatial locations. Note that in traditional Euclidean distance- and graph-based kernel design, the predictive standard deviation is equivalent to the spatial closeness between unobserved location $\vect{x}$ and the observed ones. However, to incorporate the geometric characteristics of the complex systems, our spatial kernel $\mathcal{K}_\text{s}$ is defined in the eigenspace of the manifold Laplacian, $\Delta_M$. In other words, the resulting predictive uncertainty captures the closeness between different locations in the eigenspace of $\Delta_M$, which is not the direct spatial closeness along the 3D geometry. 

In order to incorporate the direct spatial closeness of locations in the 3D geometry and avoid the selected locations that purely guided by uncertainty measure to cluster in local regions, we propose to combine the uncertainty quantification and geodesic distance-guided space-filling design to construct our active learning criterion:
    \begin{equation}
    \begin{aligned}
        \vect{x}_{n+1} = \arg\max_{\vect{x}} \bigg\{
             \alpha_{n,1} \cdot \frac{\min\limits_{i \in \{1, \dots, n\}} d_{g}\left(\vect{x}, \vect{x}_{i}\right)}{\max\limits_{\vect{x}} \min\limits_{i \in \{1, \dots, n\}} d_g\left(\vect{x}, \vect{x}_{i}\right)}
             + \alpha_{n,2} \cdot \frac{\hat{\nu}_n(\vect{x})}{\max_{\vect{x}} \hat{\nu}_n(\vect{x})}
        \bigg\}
    \end{aligned}
    \label{Eq:A-AL}
    \end{equation}
    where $d_g(\vect{x},\vect{x}_i)$ denotes the geodesic distance between points $\vect{x}$ and $\vect{x}_i$ \cite{xie2024kronecker}, $\alpha_{n,1}$ and $\alpha_{n,2}$ are weights for space-filling and uncertainty-based criteria, respectively, and the denominators serve as normalization factors.

Additionally, we propose to adaptively balance between the two criteria during the active learning process through:
    \begin{equation}
        \begin{aligned}
            \alpha_{n,1} = \frac{\hat{\tau}_{n,\text{cv}}^2}{\hat{\tau}_{n,\text{cv}}^2 + \hat{\sigma}^2_{n;\epsilon,\text{s}}},\quad 
            \alpha_{n,2} = \frac{\hat{\sigma}^2_{n;\epsilon,\text{s}}}{\hat{\tau}_{n,\text{cv}}^2 + \hat{\sigma}^2_{n;\epsilon,\text{s}}}
        \end{aligned}
        \label{eq:alpha}
    \end{equation}
    where $\hat{\sigma}^2_{n;\epsilon,\text{s}}$ is the estimated spatial noise level, and $\hat{\tau}_{n,\text{cv}}^2$ is the averaged leave-one-out cross-validation error from the G-ST-GP trained by the signals collected from $n$ spatial locations:
    \begin{equation}
        \begin{aligned}
            \hat{\tau}_{n,\text{cv}}^2 = \frac{1}{n} \sum_{i=1}^n \sum_{j=1}^{N_\text{t}} \left( y(\vect{x}_i,t_j) - \hat{\mu}_{-\vect{x}_i}(\vect{x}_i,t_j) \right)^2 \label{eq:tau_square}
        \end{aligned}
    \end{equation}
    where $y(\vect{x}_i,t_j) $ is the observed value at $(\vect{x}_i,t_j) $, and $\hat{\mu}_{-\vect{x}_i}(\vect{x}_i,t_j)$ is the corresponding prediction based on the training data excluding samples collected at spatial location $\vect{x}_i$. The adaptive weights $\alpha_{n,1}$ and $\alpha_{n,2}$ automatically update based on the goodness-of-fit of our G-ST-GP model and the spatial noise factors in each iteration. When $\hat{\tau}_{n,\text{cv}}^2$ significantly exceeds $\hat{\sigma}^2_{n;\epsilon,\text{s}}$, indicating lower reliability of the predictive model, the space-filling criterion dominates due to a larger $\alpha_{n,1}$. Conversely, when $\hat{\tau}_{n,\text{cv}}^2$ is significantly less than $\hat{\sigma}^2_{n;\epsilon,\text{s}}$, the predictive model becomes more accurate and uncertainty-based criterion takes precedence through a larger $\alpha_{n,2}$.

We observe from Eq.~\eqref{eq:tau_square} that the computation of $\hat{\tau}_{n,\text{cv}}^2$ depends on the residuals $\left(  y(\vect{x}_i,t_j) - \hat{\mu}_{-\vect{x}_i}(\vect{x}_i,t_j) \right)$ for each sample $i \in \{1, 2, \ldots, n\}$, which involves matrix inversion and multiplication for $n$ times, incurring significant computational efforts. Specifically, according to Eq. (\ref{Eq: mu}), we have
    \begin{equation}
    \begin{aligned}
        \hat{\mu}_{-\vect{x}_i}(\vect{x}_i,t_j)
        =\left (\mathcal{K}_\text{s}(\vect{x}_i,\mathcal{X}_{n(-i)})\otimes \mathcal{K}_\text{t}(t_j,\mathcal{T}_\text{tr}) \right)\left[\Sigma_{n(-i)}\otimes\Sigma_{\text{tr},\text{t}}\right]^{-1}\vect{y}_{n(-i)}
        \label{Eq: mu-i}
    \end{aligned}
    \end{equation}
    where $\mathcal{X}_{n(-i)}=\{\vect{x}_1,\dots,\vect{x}_{i-1},\vect{x}_{i+1},\dots,\vect{x}_n\}$, $\Sigma_{n(-i)}=\mathcal{K}_\text{s}(\mathcal{X}_{n(-i)},\mathcal{X}_{n(-i)})+\hat{\sigma}_{n;\epsilon,s}I_{n-1}$ and $\vect{y}_{n(-i)}=[y(\vect{x}_k,t_j)]_{\vect{x}_k\in \mathcal{X}_{n(-i)},t_j\in\mathcal{T}_\text{tr}}$. Thus, we need to design a computationally efficient way to calculate $\hat{\tau}_{n,\text{cv}}^2$.

If we define $\Sigma_{n(-i),N_\text{t}}=\Sigma_{n(-i)}\otimes\Sigma_{\text{tr},\text{t}}$ and $\Sigma_{n,N_\text{t}}=(\mathcal{K}_s(\mathcal{X}_n,\mathcal{X}_n)+\hat{\sigma}_{n;\epsilon,s}I_{n})\otimes \Sigma_{\text{tr},\text{t}}$, to isolate the dynamic signals collected from the $i^\text{th}$ spatial location, we partition $\Sigma_{n,N_\text{t}}$ into block matrices:
    \begin{equation}
        \Sigma_{n,N_\text{t}} = \begin{pmatrix}
                \Sigma_{n(-i),N_\text{t}} & \mathcal{K}_{(1_i, n - 1_i),N_\text{t}}^\top \\
                \mathcal{K}_{(1_i, n - 1_i),N_\text{t}} & \mathcal{K}_{(1_i, 1_i),N_\text{t}}
            \end{pmatrix}
        \label{eq:sigma_block}
    \end{equation}
    where $\mathcal{K}_{(1_i, n - 1_i),N_\text{t}} = \mathcal{K}_\text{s}(\vect{x}_i,\mathcal{X}_{n(-i)}) \otimes \mathcal{K}_\text{t}(\mathcal{T}_\text{tr},\mathcal{T}_\text{tr})$ is the covariance between the dynamics at the $i^\text{th}$ location and the rest of the training data, and $\mathcal{K}_{(1_i, 1_i),N_\text{t}} = \left(\mathcal{K}_\text{s}(\vect{x}_i,\vect{x}_i)+\hat{\sigma}_{n;\epsilon,\text{s}}\right) \otimes \Sigma_{\text{tr},\text{t}}$ is the covariance matrix for the dynamics at the $i^\text{th}$ location. Using the block matrix inversion formula, we have:
    \begin{equation}
        \begin{aligned}
            \Sigma_{n,N_\text{t}} ^{-1} &:= \begin{pmatrix}
                A_{11} & A_{21}^\top \\
                A_{21} & A_{22}
            \end{pmatrix}        
        \end{aligned}
        \label{eq:inv-block}
    \end{equation}
    where $A_{21} = -A_{22} \mathcal{K}_{(1_i, n - 1_i),N_\text{t}} \Sigma_{n(-i),N_\text{t}}^{-1}$ and $A_{22} = \left( \Sigma_{n,N_\text{t}}^{-1} \right)_{(1_i, 1_i)N_\text{t}}$ is the sub-matrix of $\Sigma_{n,N_\text{t}}^{-1}$ corresponding to the $i$-th row of the spatial kernel matrix coupled with the temporal kernel matrix.

Hence, we can simplify Eq.~\eqref{Eq: mu-i} and calculate the residual as:
    \begin{equation}
        \begin{aligned}
            \left[y(\vect{x}_i,t_j) - \hat{\mu}_{-\vect{x}_i}(\vect{x}_i,t_j)\right]_{t_j\in \mathcal{T}_\text{tr}}
            &= A_{22}^{-1} \left( A_{22} \vect{y}_{(i)} -A_{22} \mathcal{K}_{(1_i, n - 1_i),N_\text{t}} \Sigma_{n(-i),N_\text{t}}^{-1}  \vect{y}_{n(-i)} \right)\\
            &= A_{22}^{-1} \left( A_{22} \vect{y}_{(i)} + A_{21} \vect{y}_{n(-i)} \right) \\
            &= \left( \left( \Sigma_{n,N_\text{t}}^{-1} \right)_{(1_i, 1_i)N_\text{t}} \right)^{-1} \left( \Sigma_{n,N_\text{t}}^{-1} \mathbf{y}_{n} \right)_{(1_i, \cdot)N_\text{t}}
        \end{aligned}
        \label{eq:residual_implified}
    \end{equation}
    where $\vect{y}_{(i)}=\left[y(\vect{x}_i,t_j)\right]_{t_j\in \mathcal{T}_\text{tr}} $, $\vect{y}_n=[\vect{y}_{(i)},\vect{y}_{n(-i)}]$, and $\left( \Sigma_{n,N_\text{t}}^{-1} \mathbf{y}_{n} \right)_{(1_i, \cdot)N_\text{t}}$ denotes the sub-vector of $ \Sigma_{n,N_\text{t}}^{-1} \mathbf{y}_{n} $ corresponding to the $i$-th spatial location. As such, we only need to compute the inverse of $\Sigma_{n,N_\text{t}}$ once (which can be efficiently done by exploiting the Kronecker properties) to compute the residuals at all the $n$ locations and further calculate the leave-one-out cross-validation error $\hat{\tau}_{n,\text{cv}}^2$ for each active learning step. This enables the computation-efficient implementation of our A-AL procedure for effective sequential data collection.

  \section{Numerical Experiments} \label{s:results}
 
We assess the performance of our G-ST-GP framework in predictive modeling of electrodynamics within a 3D heart geometry under ventricular fibrillation. The geometry is discretized into 1,094 nodes and 2,184 mesh elements, constituting a refined mesh derived from the geometry data in the 2007 PhysioNet Computing in Cardiology Challenge~\cite{goldberger2000physiobank}. The dynamic signals under the fibrillation condition are generated by simulating a regularly paced activation source at the apex of the geometry and numerically solving the Aliev Panfilov model using finite element methods~\cite{trayanova2011whole,clayton2008guide,aliev1996simple}. Additionally, an electrical impulse source on the right ventricle is modeled to fire more frequently than the regular trigger. The electrical signals generated by the two triggers will interact with each other, leading to chaotic electrodynamics and further imitating cardiac activities under the fibrillation condition~\cite{haissaguerre1998spontaneous,ten2009organization}. We denote the resulting simulation data as $\vect{u}(\vect{x}, t)=[u(\vect{x}_i, t_j)]_{i\in\{1,\dots,1094\},j\in\{1,\dots, 1981\}}$. Note that the time series signals collected at each spatial location $\vect{x}_i$ consist of 1981 data points, i.e., $N_t=1981$. Measurement noise is inevitable in real-world data collection. Thus, we add different levels of noise to the simulation data $\vect{u}(\vect{x}, t)$ to investigate the prediction performance. Specifically, the physical measurements are generated as
    $
    \vect{y}(\vect{x}, t) = \vect{u}(\vect{x}, t) + \xi(\vect{x}, t)
    $
    where $\xi(\vect{x}, t)$ is the measurement noise that follows a Gaussian distribution, $\xi(\vect{x}, t) \sim \sigma_\xi \cdot \mathcal{N}(0, 1)$, where $\sigma_\xi$ is the noise level coefficient.
 
The prediction performance of our G-ST-GP is compared with traditional approaches lacking the mechanism of geometric information incorporation or effective data collection. Specifically, prediction performance is compared with the popular Euclidean distance-based GP (E-ST-GP), and the efficacy of our A-AL criterion is benchmarked against non-informative random active learning (R-AL), pure uncertainty-based search (U-AL) and pure space-filling strategy (S-AL). Furthermore, to demonstrate the advantage of the adaptive method, we also compare a series of fixed combined criteria (F-AL), where the weights, i.e., $\alpha_{n,1}$ and $\alpha_{n,2}$, for both the prediction uncertainty and the space-filling design are constants. The prediction performance is evaluated based on the relative error ($RE$):
    \begin{align}
        RE &= \frac{\| \vect{\hat{u}}(\vect{x}, t) - \vect{u}(\vect{x}, t) \|}{ \| \vect{u}(\vect{x}, t) \|}
    \end{align}
    where $\vect{u}(\vect{x}, t)$ and $\vect{\hat{u}}(\vect{x}, t)$ denote the reference and predicted cardiac dynamics, respectively. 

\begin{figure}
	\centering
	\includegraphics[width=0.6\linewidth]{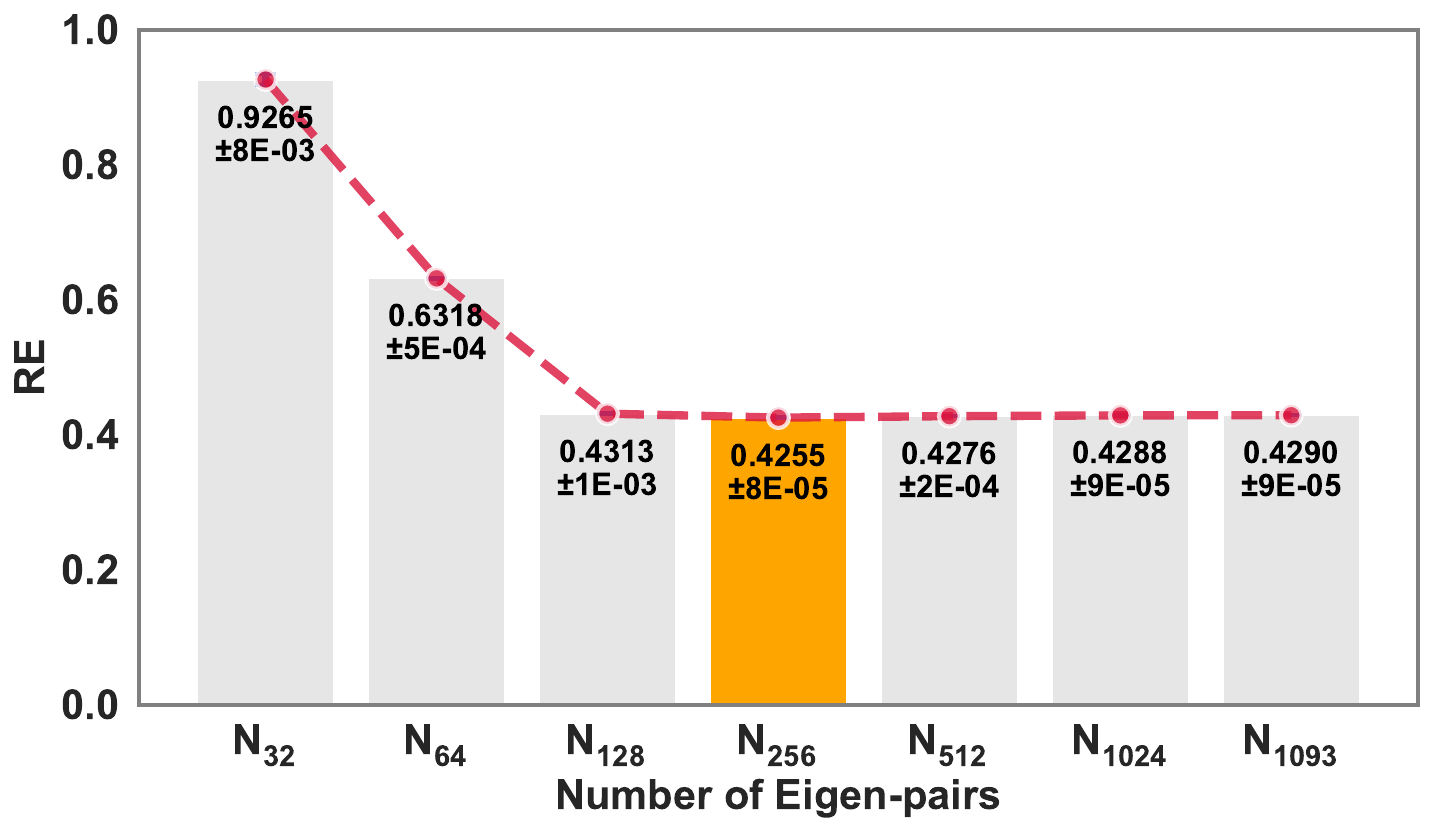}
    \caption{The variation of $RE$ with the number of eigenpairs. %The optimal eigen-pair number is highlighted at $N_{256}$, i.e., with $256$ eigen-pairs.
    }
	\label{Fig:Pars}
\end{figure}
 
\subsection{\emph{Prediction of Cardiac Electrodynamics}}
In our G-ST-GP model, one crucial hyper-parameter is the number of eigenpairs used to construct the spatial kernel (see Eq. (\ref{eq: def-kernel-spa})). Fig. \ref{Fig:Pars} illustrates the variation of $RE$ yielded by the G-ST-GP with different numbers of eigenpairs trained by data collected from 50 spatial locations, i.e., $\mathcal{X}_{tr}=\{\vect{x}_1,\dots,\vect{x}_{50}\}$. Note that $RE$ decreases sharply from 0.9265 at $N_{32}$ to 0.4313 at $N_{128}$, followed by a stabilization phase with the $RE$ values ranging from 0.4255 to 0.4290. Particularly with 256 eigenpairs, i.e., $N_{256}$, the $RE$ reaches the minimum value of 0.4255, suggesting a balance between prediction accuracy and computational efficiency. As such, we use 256 eigenpairs to construct the Laplacian-based spatial kernel.

\begin{figure*}
	\centering
	\includegraphics[width=5.5in]{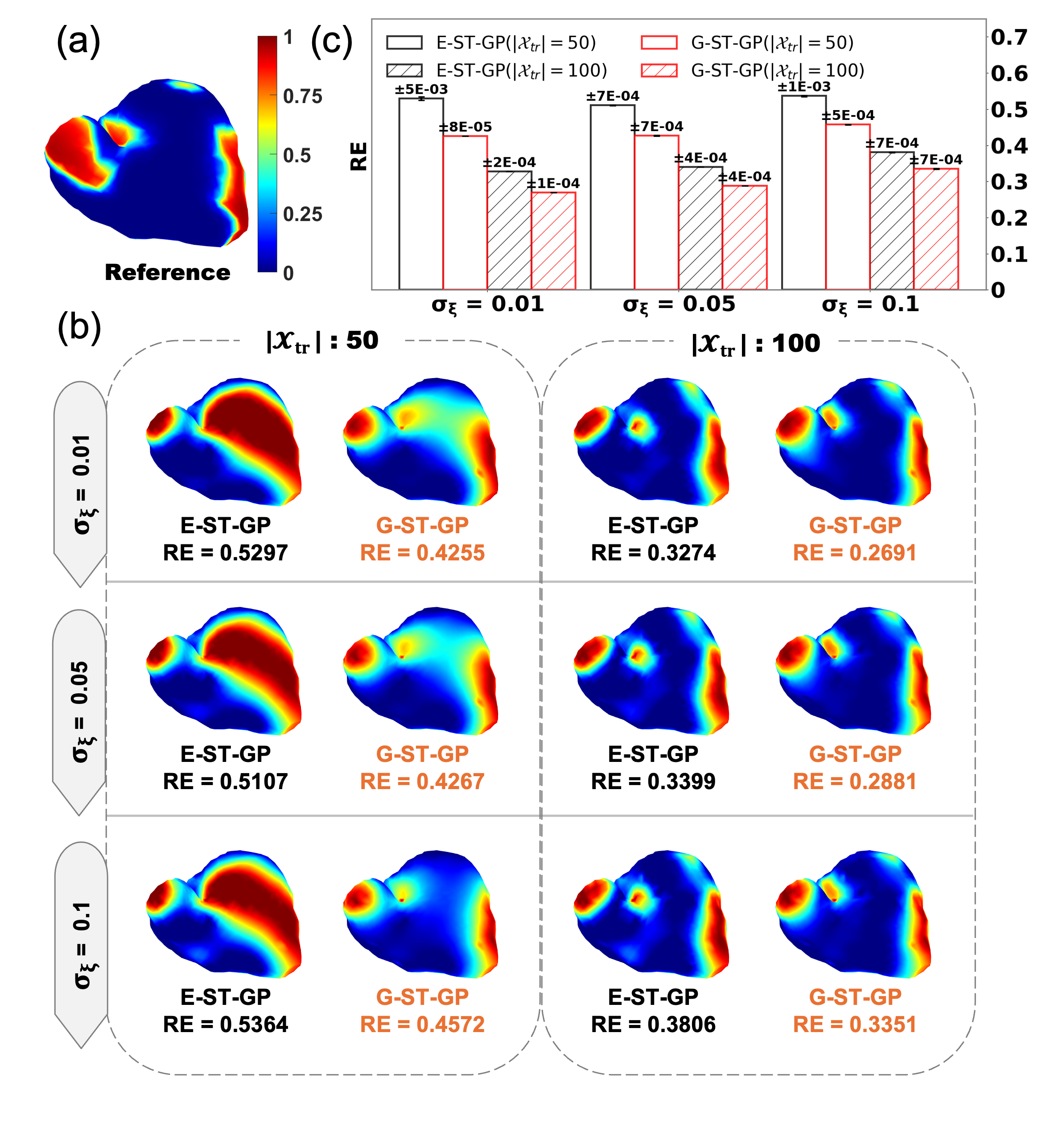}
	\caption{(a) Reference mapping of the electrodynamics at time point $t=1550$. (b) Estimated mappings produced by E-ST-GP and G-ST-GP under different training dataset sizes ($|\mathcal{X}_\text{tr}| = 50$ or $100$) and noise levels ($\sigma_\xi = 0.01, 0.05, 0.1$) at time $t=1550$. (c) Bar chart comparing the average $RE$'s for E-ST-GP and G-ST-GP from 5 replications across different scenarios in (b).}
	\label{Fig:Prediction}
\end{figure*}

Fig.~\ref{Fig:Prediction} presents a comparative analysis of prediction performance between our G-ST-GP and the traditional E-ST-GP, evaluated across two training set sizes and three noise levels. Fig.~\ref{Fig:Prediction}(a) illustrates the ground truth distribution of the signals on the heart surface. The spatial distribution of electrodynamics, which evolves dynamically over time, is visualized at time step $t = 1550$ out of $1981$ total steps. Fig.~\ref{Fig:Prediction}(b) displays the estimated cardiac electrodynamics mappings generated by both methods under varying conditions: training dataset sizes of $|\mathcal{X}_\text{tr}| = 50$ and $100$, and noise levels of $\sigma_\xi = 0.01$, $0.05$, and $0.1$. Our G-ST-GP demonstrates superior performance in preserving spatial patterns of the reference mapping shown in Fig.~\ref{Fig:Prediction}(a), while the E-ST-GP predictions exhibit more significant deviations across all noise levels. This enhanced predictive capability of the G-ST-GP model can be attributed to its effective incorporation of both temporal correlations and geometric manifold features, enabling robust predictions of spatiotemporal cardiac electrodynamics.

Fig. \ref{Fig:Prediction}(c) provides a bar chart comparing the $RE$'s generated by both the G-ST-GP and E-ST-GP methods under different situations. Note that our G-ST-GP method consistently yields a smaller $RE$ compared to E-ST-GP, regardless of the training dataset size or noise level. Specifically, under the noise condition of $\sigma_{\xi}=0.01$, the $RE$ yielded by G-ST-GP is reduced by $19.67\%$ and $17.81\%$ for $|\mathcal{X}_\text{tr}|=50$ and $|\mathcal{X}_\text{tr}|=100$ compared to E-ST-GP, respectively. Similarly, reductions of $16.55\%$ and $15.26\%$ in $RE$ by G-ST-GP are observed for $|\mathcal{X}_\text{tr}|=50$ and $|\mathcal{X}_\text{tr}|=100$ respectively, when $\sigma_{\xi}=0.05$. When the noise level further increases to $\sigma_\xi=0.1$, our G-ST-GP method achieves a reduction in $RE$ of $14.77\%$ and $11.97\%$ for $|\mathcal{X}_\text{tr}|=50$ and $|\mathcal{X}_\text{tr}|=100$, respectively. It is also worth noting that the $RE$ generated by the G-ST-GP method significantly improves when the size of spatial measurement locations increases from 50 to 100, achieving an improvement of $36.76\%$, $32.48\%$, and $26.71\%$ under the noise condition of $\sigma_{\xi}=0.01$, $\sigma_{\xi}=0.05$, and $\sigma_{\xi}=0.1$, respectively. This observation demonstrates that the quality and volume of training data significantly impact the predictive power of the G-ST-GP model.

\subsection{\emph{Adaptive Active Learning}}

\begin{figure*}
    \centering
    \includegraphics[width=6.0in]{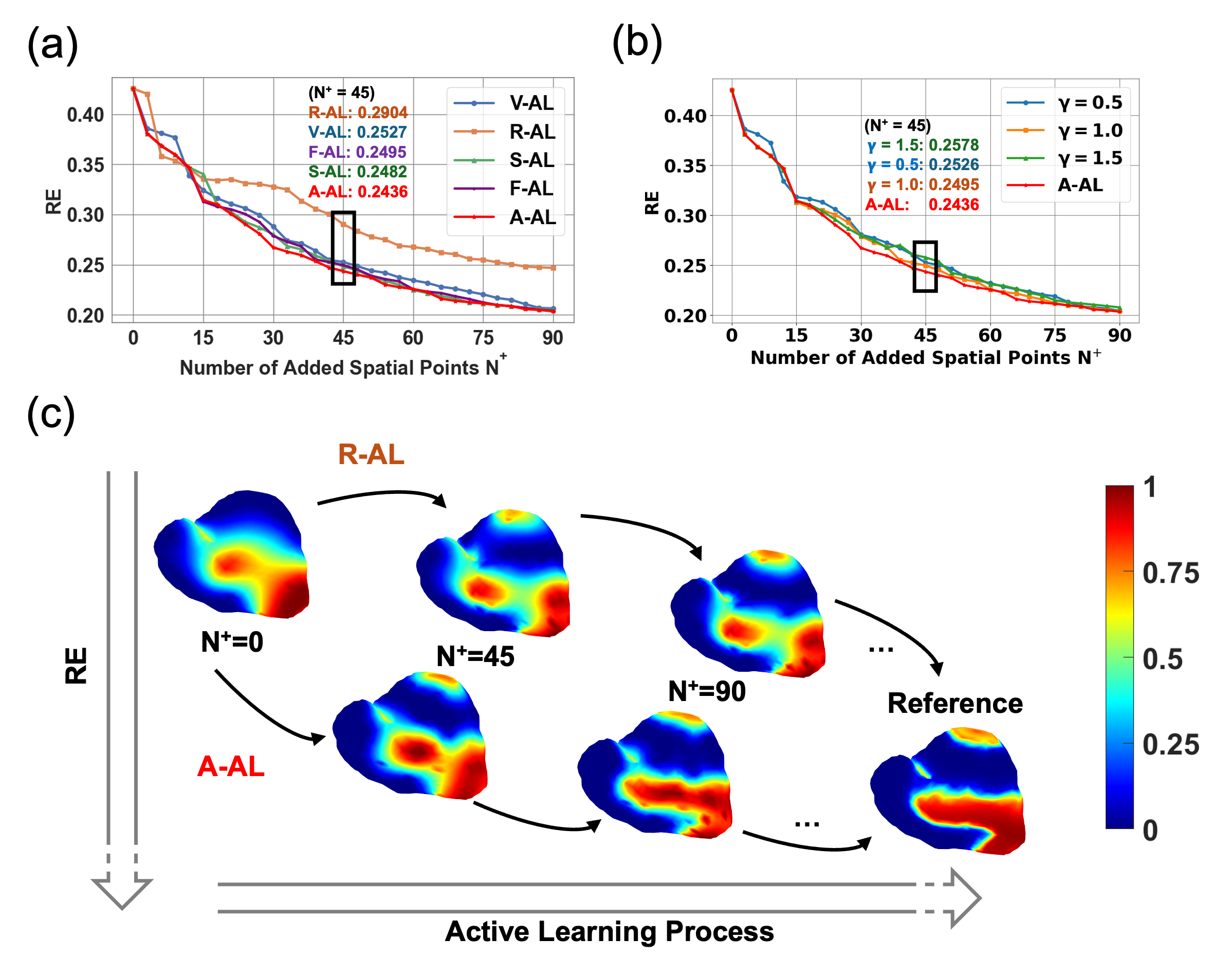}
    \caption{(a) Comparison of the G-ST-GP model performance trained by signals collected from our A-AL strategy versus other methods including V-AL, S-AL, and F-AL with $\alpha_{n,1}=\alpha_{n,2}=0.5$ when $\sigma_\xi = 0.01$. (b) Comparison of A-AL and different F-AL approaches with $\gamma = \alpha_{n,1}/\alpha_{n,2}=0.5,\ 1.0$ and 1.5. (c) Performance evolution of the active learning processes provided by A-AL and R-AL (at time point $t=730$), shown alongside the initial estimation at $N^{+}=0$ and ground truth for reference.}
    \label{Fig:AL}
\end{figure*}

To assess the effectiveness of our proposed A-AL strategy, we further conduct a series of experiments. The initial G-ST-GP model is trained by signals collected from randomly sampled 50 spatial locations. With the well-trained initial model, we compute the posterior standard deviation $\hat{\nu}(\vect{x})$ for each unmeasured location across the heart geometry. This uncertainty quantification is subsequently integrated with a geodesic-distance-based space-filling design to evaluate our A-AL criterion (see Eq.~\eqref{Eq:A-AL}), enabling the identification of sub-sequent query points for physical observations. The weights to balance these two criteria are automatically calculated following Eq. (\ref{eq:alpha}). To optimize computational efficiency, each A-AL iteration selects three spatial locations for subsequent sensor deployment. Note that each selected location will be removed from the candidate pool of unmeasured points and will be further incorporated into the set of existing locations for future A-AL evaluations. We repeated this iterative process for 30 rounds to ensure robust validation of our approach.

Fig. \ref{Fig:AL}(a) compares the G-ST-GP model performance trained using data collected from our A-AL procedure with traditional methods including V-AL, S-AL, and the method that combines uncertainty quantification and space-filling design but with fixed weights (F-AL) of $\alpha_{n,1}=\alpha_{n,2}=0.5$ when $ \sigma_\xi = 0.01 $. Overall, the curve for our A-AL method remains as the lower bound in most rounds among the five methods. When 45 new spatial locations (i.e., $N^+=45$) are added to the observation set, the $RE$ is reduced to 0.2436 by A-AL, achieving the minimal value compared with F-AL (0.2495), S-AL (0.2482), V-AL (0.2527) and R-AL (0.2904). When $N^+$ further reaches 90, the $RE$ provided by A-AL, F-AL, S-AL, and V-AL converges to a similar value. This might be due to the fact that the information gain given by additional observations saturates for G-ST-GP modeling. Nonetheless, our A-AL procedure achieves the smallest $RE$ of 0.2036 compared with F-AL (0.2042), S-AL (0.2048) and V-AL (0.2064). Additionally, A-AL achieves a significant reduction in $RE$ of 17.54\% compared with R-AL (0.2469). As such, our A-AL-based G-ST-GP is able to reliably predict the electrodynamics over the entire geometry given sensor observations from only about 10\% (140/1094) of the spatial locations.

To further assess the advantage of our A-AL compared with empirical weight-setting methods, i.e., F-AL, we conduct additional experiments with different fixed weight settings. Fig. \ref{Fig:AL}(b) shows the comparison of our A-AL strategy and the F-AL method with fixed weight ratios of $\gamma= \alpha_{n,1}/\alpha_{n,2} =0.5, 1.0$ and $1.5$. Our A-AL method generally maintains a lower $RE$ compared with the F-AL methods. Specifically, at $N^{+}=45$, A-AL achieves a smaller $RE$ of 0.2436, compared with F-AL that yields an $RE$ of 0.2526 ($\gamma=0.5$), 0.2495 ($\gamma=1.0$) and 0.2578 ($\gamma=1.5$). When the total number of the added spatial locations reaches 90, the $RE$ further decreases to 0.2036 for A-AL compared with the $RE$ of F-AL under $\gamma=0.5$ (0.2045), $\gamma=1.0$ (0.2042) and $\gamma=1.5$ (0.2079).

Fig. \ref{Fig:AL}(c) illustrates the evolution of the dynamic mappings during the active learning processes provided by our A-AL strategy and the non-informative R-AL approach. Both methods exhibit noticeable improvements in the accuracy of the estimated mappings when the number of additional locations for data collection to train the G-ST-GP model increases from $N^{+}=0$ to $N^{+}=45$ and $N^{+}=90$. The visualization also highlights the clear advantage of the A-AL strategy, which significantly outperforms R-AL in predicting cardiac electrodynamics. This enhanced performance can be attributed to the more efficient selection of sensor measurements under the A-AL framework, resulting in a higher concentration of informative data points that contribute significantly to the model's accuracy.

  \section{Conclusions} \label{s:conclusions}

In this paper, we propose an effective geometry-aware active learning framework via Gaussian Process for predictive modeling of spatiotemporal dynamic systems. We develop a geometry-aware spatiotemporal Gaussian Process (G-ST-GP) that integrates temporal correlations with geometric manifold features to enhance the prediction of high-dimensional dynamic behaviors. Furthermore, we introduce an adaptive active learning (A-AL) strategy that combines prediction uncertainty in G-ST-GP with a geodesic distance-guided space-filling scheme to identify informative spatial locations for data collection, thereby enhancing the model's prediction performance. Additionally, we exploit the properties of Kronecker product of the spatiotemporal kernel in G-ST-GP to improve computational efficiency in the posterior modeling and adaptive active learning process. We validate our framework through numerical studies in 3D spatiotemporal cardiac modeling. Numerical results demonstrate that our approach significantly outperforms traditional methods that are often limited by insufficient geometric information incorporation or suboptimal training data collection. These findings highlight the potential of our framework to advance the predictive modeling of complex spatiotemporal systems, offering a powerful tool for tackling real-world challenges in dynamic system analysis.     

  \section{Data Availability Statement}
    The mesh and geometry data that support the findings of this study are openly available in the PhysioNet/Computing in Cardiology Challenge 2007 at \url{https://physionet.org/content/challenge-2007/1.0.0/}, reference number ~\cite{goldberger2000physiobank}.

  \section*{Acknowledgement}
 % \spacingset{1}
  This research work was supported by the National Heart, Lung, And Blood Institute of the National Institutes of Health under Award Number R01HL172292. The content is solely the responsibility of the authors and does not necessarily represent the official views of the National Institutes of Health. 

  {
    \bibliographystyle{apalike}
    \spacingset{1}
    \bibliography{Ref}
  }

\end{document}